\documentclass[a4paper,fleqn]{cas-dc}

\usepackage[numbers]{natbib}

\usepackage{graphicx}
\usepackage{multirow}
\usepackage{tabularray}
\usepackage{caption}
\usepackage{subcaption}
\usepackage[linesnumbered,ruled,vlined]{algorithm2e}
\usepackage{xcolor}

\def\tsc#1{\csdef{#1}{\textsc{\lowercase{#1}}\xspace}}
\tsc{WGM}
\tsc{QE}

\begin{document}
\let\WriteBookmarks\relax
\def\floatpagepagefraction{1}
\def\textpagefraction{.001}

\shorttitle{FedTVD}

\shortauthors{}

\title [mode = title]{FedTVD: Balancing Data Quality and Quantity for Robust Federated Learning}



%

\author{Radwan Selo}



\ead{radwan-selo@cbnu.ac.kr}

\credit{Conceptualization, Methodology, Visualization, and Writing – original draft}

\author{Majid Kundroo}
\ead{kundroomajid@cbnu.ac.kr}

\credit{Formal analysis, Validation, and Writing – review and editing}

\affiliation{organization={School of Information and Communication Engineering, Chungbuk National University},
            city={Cheongju},
            postcode={28644}, 
            country={Republic of Korea}}

\author{Taehong Kim}

\cormark[1]


\ead{taehongkim@cbnu.ac.kr}


\credit{Resources, Validation, Supervision and Writing – review and editing}


\cortext[1]{Corresponding author}


\nonumnote{
\textcopyright~2026.~This~manuscript~version~is~made~available~under~the~CC-BY-NC-ND~4.0~license~(\url{http://creativecommons.org/licenses/by-nc-nd/4.0/}).~This~is~the~accepted~manuscript~version~of~an~article~published~in~\textit{Future~Generation~Computer~Systems}.~The~final~published~version~is~available~at~\url{https://doi.org/10.1016/j.future.2025.108177}.
}

\begin{abstract}
Federated Learning (FL) enables collaborative model training across distributed client devices while preserving data privacy. However, FL faces significant challenges due to data heterogeneity, particularly in terms of label distribution skewness and variations in dataset sizes, which can lead to biased model updates and hinder convergence. To address this, we propose FedTVD, a novel FL algorithm that weights client contributions during aggregation by considering both data quality and quantity. Unlike traditional FL approaches such as FedAvg, which rely solely on dataset size for client weighting, FedTVD integrates Total Variation Distance (TVD) to measure the divergence between each client’s local label distribution and a uniform global distribution. Clients with highly skewed distributions receive lower weights, preventing unbalanced datasets with imbalances from disproportionately influencing the global model.
At the same time, dataset size is incorporated to ensure scalability and fairness. This dual-weighting mechanism effectively mitigates the impact of data imbalance, leading to more stable and generalized global models. Experimental results show that FedTVD consistently outperforms state-of-the-art methods across all datasets (FMNIST, CIFAR-10, and CIFAR-100) and all levels of data heterogeneity. Notably, it achieves up to 10.6\% improvement over FedAvg on CIFAR-10 under highly skewed data, while maintaining top performance even under moderate and IID settings.
\end{abstract}



\maketitle

\section{Introduction}\label{sec1}

FL \cite{konečný2017federatedlearningstrategiesimproving} has emerged as a revolutionary paradigm in distributed machine learning, enabling collaborative model training across multiple decentralized clients without requiring them to share their raw data. This design inherently enhances data privacy and security, addressing significant concerns in sensitive domains such as healthcare, finance, and personalized devices \cite{yang2019federated, LI2020106854, ZHANG2021106775}. In contrast to traditional centralized learning, where data is aggregated in a single location, FL leverages local training on client devices, transmitting only model updates to a central server for aggregation. This decentralized approach aligns with stringent data privacy regulations, such as the General Data Protection Regulation (GDPR), and mitigates the risks associated with transferring large volumes of sensitive data over networks \cite{kairouz2021advancesopenproblemsfederated, XU2024146}. Moreover, FL significantly reduces communication overhead and computational burden on centralized systems, making it well-suited for large-scale, heterogeneous environments.

Despite its advantages, FL faces several challenges that hinder its practical deployment in real-world scenarios. These challenges can be broadly categorized into communication efficiency, system heterogeneity, and data heterogeneity. Communication efficiency is critical as FL systems often operate in resource-constrained environments, such as edge devices or Internet of Things (IoT) networks, where bandwidth and latency \cite{10681540, aedfl} are limiting factors \cite{10086519, BARRETO2025107514, APAT2023100866, computers14030099}. System heterogeneity, on the other hand, arises from the diverse computational capacities of client devices, which can lead to unequal participation and delays in model updates \cite{10272562}. Among these, data heterogeneity, where the data across clients is non-Independent and Identically Distributed (non-IID) exhibit significant variations in size and class distributions, hence. This issue is inherently tied to the decentralized nature of the system. Clients typically generate data based on their local environments and user behaviors \cite{li2019convergence}. For example, healthcare organizations contributing data to a federated medical imaging model may have patient populations that differ in demographics, disease prevalence, and imaging protocols, while mobile devices participating in FL-based recommendation systems may exhibit distinct usage patterns. This heterogeneity becomes particularly problematic during model aggregation \cite{zhao2018federatedLink, ZHU2021371}, as some clients possess large but skewed datasets, while others contribute smaller yet more balanced datasets. Standard aggregation methods, such as FedAvg \cite{mcmahan2017communication}, tend to prioritize clients with larger datasets, often resulting in global models that overfit dominant distributions \cite{Hu_2023} and struggle with fairness and generalization. Addressing this imbalance is crucial for developing robust and equitable FL models that generalize well across diverse client populations.

A key barrier to effective FL is ensuring that the global model remains both fair and robust despite these challenges. This challenge is amplified by the fact that clients often learn from skewed local data, while the global model must perform well on more balanced data. This mismatch highlights the importance of a client’s data quality, particularly its label distribution. However, considering only quality may undervalue clients with large but imbalanced data. Existing aggregation strategies often lead to models that overfit to dominant data distributions, causing model drift \cite{jothimurugesan2023federatedlearningdistributedconcept, modeldrift2}, a phenomenon where the model's performance deteriorates as local data deviates from the global model over time. This drift is particularly prevalent with non-IID dataset, where clients hold data with varying class distributions. Thus, it is essential to consider both the distributional quality and quantity of each client’s data, penalizing highly skewed distributions while still leveraging useful large-scale data, to support stable convergence and better generalization. To address these challenges, this study introduces \emph{Federated Learning with Total Variation Distance (FedTVD)}, a novel algorithm designed to mitigate the impact of heterogeneous label distributions. By leveraging the TVD \cite[Chapter 4, p. 47]{levin2017markov}, a robust and interpretable metric for quantifying the divergence between probability distributions, FedTVD employs a dual weighting mechanism that balances label distribution skewness with dataset size, thereby promoting better convergence, improved generalization, and enhanced fairness. By intelligently adjusting each client’s influence based on the nature of its data, our approach aims to ensure that the final model is both inclusive and robust. Ultimately, by addressing the inherent data heterogeneity among clients, we aim to create a system that achieves a balanced integration of diverse data sources, leading to a global model that is more accurate, generalizable, and fair.

In light of the challenges posed by data heterogeneity and the limitations of traditional aggregation methods, this work makes several significant contributions to advancing FL. In particular, our work provides a comprehensive algorithm that not only addresses fairness and robustness in global model training but also enhances overall performance through practical, scalable solutions. Our key contributions are as follows:
\begin{enumerate}
    \item \textbf{TVD-based Weighting:} We introduce a novel weighting mechanism that assigns adaptive weights to clients based on both label distribution skewness and dataset size. This ensures that clients with heavily skewed data have a moderated influence on the global model, while those with substantial and balanced data are appropriately emphasized, thereby promoting better convergence, improved generalization, and enhanced fairness.
    \item \textbf{Empirical Validation:} Through extensive empirical evaluations across diverse datasets and application scenarios, we demonstrate that FedTVD consistently outperforms existing methods, particularly in environments with highly heterogeneous data. By dynamically adjusting client contributions based on data quality and quantity, FedTVD ensures stable convergence and enhanced performance, making it particularly suitable for large-scale, non-IID environments.
\end{enumerate}

The remainder of this paper is structured as follows: Section \ref{sec2} reviews related work in FL, exploring various methods used to handle challenges like non-IID data and model drift. Section \ref{sec3} describes the methodology of this study, outlining the problem formulation, including model drift and data quality/quantity, and detailing the innovative TVD-based weighting mechanism used to improve global model convergence. Section \ref{sec4} explains the experimental setup, providing information on the environment, dataset distribution strategies, hardware configuration, model architecture and hyperparameters employed in the study. Section \ref{sec5} presents the results and a comprehensive discussion, highlighting the performance of the FedTVD approach under both IID and non-IID data conditions. Additionally, experiments were conducted to evaluate the effect of increasing the number of clients and the Client Participation Rate (CPR) on model accuracy. Finally, Section \ref{sec6} concludes the paper, summarizing the key findings and suggesting directions for future research.

\section{Related Works}\label{sec2}

\begin{table*}
\centering
\caption{Summary of Key FL Methods, Addressed Challenges and Limitations}
\label{tab:relatedworksummary}
\begin{tblr}{
  width = \linewidth,
  colspec = {Q[160]Q[40]Q[208]Q[144]Q[237]},
  column{2} = {c},
  hlines,
  vlines,
}
\textbf{Method} & \textbf{Year} & \textbf{Key Idea} & \textbf{Addressed Challenges} & \textbf{Limitations}\\
FedAvg \cite{mcmahan2017communication} & 2017 & Weighted average of local models by dataset size & Data quantity imbalance (assumes IID data) & Poor performance with non-IID data bias toward majority distributions\\
FedProx \cite{li2020federatedoptimizationheterogeneousnetworks} & 2020 & Adds proximal term to local objectives for stability & System heterogeneity & Insufficient for distributional (label/feature) heterogeneity\\
FedNova \cite{wang2020tacklingobjectiveinconsistencyproblem} & 2020 & Normalizes client updates by training progress & Computational (system) imbalance & Assumes data distribution uniformity poor with skewed data\\
FedMA \cite{wang2020federatedlearningmatchedaveraging} & 2020 & Aligns and averages layer-wise model representations across clients & Feature representation heterogeneity & Does not explicitly account for variations in class distribution skew or disparities in client dataset sizes during aggregation\\
FedBN \cite{li2021fedbnfederatedlearningnoniid} & 2021 & Local batch normalization parameters global weight aggregation & Feature distribution shift & Does not address label skew or data quantity imbalance\\
Scaffold \cite{karimireddy2021scaffoldstochasticcontrolledaveraging} & 2021 & Control variates to reduce client drift and update divergence & Client drift and training stability & Requires careful tuning limited handling of label and feature imbalance\\
FedDkw \cite{feddkw} & 2023 & Weights updates by KL divergence between client and global data distributions & Data distribution heterogeneity & Ignores dataset size and unbounded KL may destabilize aggregation\\
BN-Scaffold \cite{quintana2024bnscaffoldcontrollingdriftbatch} & 2024 & Incorporates local batch normalization into Scaffold to address drift and feature shift & Client drift and feature shift & May depend on hyperparameter tuning does not fully resolve label skew or dataset size imbalance\\
Clustered FL \cite{sattler2019clusteredfederatedlearningmodelagnostic,li2024federatedlearningclientsclustering,DENG2024835} & 2021-2024 & Client clustering to train subgroup-specific models & Highly diverse client data distributions & Relies on accurate and stable clustering struggles with dynamic client populations\\
Personalized FL \cite{9743558,Zhang_Hua_Wang_Song_Xue_Ma_Guan_2023} & 2023 & Client-specific model adaptation or fine-tuning of global model & Local data heterogeneity and personalization & Adds complexity focuses on local optimization not global solutions\\
FedOpt \cite{reddi2021adaptivefederatedoptimization} & 2021 & Adaptive optimizers applied to global aggregation & Convergence under data and system heterogeneity & Depend on hyperparameter
tuning and does not fully resolve data skew
\end{tblr}
\end{table*}

FL has been the focus of extensive research, with numerous algorithms proposed to address its inherent challenges. A key area of exploration is mitigating the effects of non-IID data across clients, which significantly impact the efficiency and fairness of collaborative training. This section reviews prominent FL algorithms and their strategies for handling data heterogeneity, highlighting their contributions and limitations in the context of developing more robust aggregation methods, with a summarized overview provided in Table \ref{tab:relatedworksummary}.

\subsection{Data Distribution Heterogeneity}

Balancing the contribution of clients with diverse dataset sizes and data distributions remains a fundamental challenge in FL. Classical methods such as FedAvg aggregate updates using dataset-size-weighted averaging, which is effective for IID data but can result in biased models under non-IID scenarios. FedProx \cite{li2020federatedoptimizationheterogeneousnetworks} augments this by penalizing local updates that deviate from the global model, leading to greater algorithmic stability. However, neither FedAvg nor FedProx fully mitigates issues arising from non-uniform label or feature distributions. FedNova \cite{wang2020tacklingobjectiveinconsistencyproblem} further adjusts for system heterogeneity by normalizing updates based on each client's local progress. More recently, FedDkw \cite{feddkw} has introduced KL divergence to weight the similarity between client and server label distributions for aggregation, yet this approach can suffer from instability due to the unbounded nature of KL and neglects the size of each client’s dataset. Overall, while these algorithms address parts of the problem, a unified solution that robustly balances both data quantity and distributional diversity is still lacking.

\subsection{Feature Heterogeneity}

Distinct feature distributions among clients are especially prominent in domains such as medical imaging and multi-institutional datasets. FedBN \cite{li2021fedbnfederatedlearningnoniid} provides an effective remedy by maintaining client-specific batch normalization parameters while synchronizing only the remaining model weights. This preserves locality in feature statistics, improving global performance where input distributions vary. Another approach, FedMA \cite{wang2020federatedlearningmatchedaveraging}, aligns and averages layer-wise model representations across clients to reduce feature representation mismatch during aggregation. These strategies enhance model robustness under heterogeneous input spaces, but they still largely overlook skew in class distributions and disparities in dataset sizes.

\subsection{Client Drift}

Local model updates often diverge from the global trajectory in the presence of data heterogeneity, resulting in diminished convergence rates. Scaffold \cite{karimireddy2021scaffoldstochasticcontrolledaveraging} directly addresses this challenge by utilizing control variates, which help align local updates with the direction of global optimization and significantly improve training stability under non-IID data. A recent variant, BN-Scaffold \cite{quintana2024bnscaffoldcontrollingdriftbatch}, enhances this by incorporating local batch normalization into the Scaffold framework, thereby addressing both update divergence and feature shift in heterogeneous environments. Although these approaches demonstrate improved convergence, they do not fully resolve underlying label skew or dataset size imbalance, and their effectiveness may depend on careful hyperparameter tuning.

\subsection{Clustered Federated Learning}

Clustered FL addresses the challenge of highly diverse data distributions by partitioning clients into subgroups with similar local data characteristics, enabling each cluster to train its own specialized model \cite{li2024federatedlearningclientsclustering,DENG2024835,sattler2019clusteredfederatedlearningmodelagnostic, fldqn}. These clusters are typically formed using gradient-based statistics, distributional similarity, or auxiliary metadata. By aligning models more closely with subgroup distributions, clustered FL improves personalization and convergence in non-IID settings. However, the effectiveness of this approach relies heavily on accurate and stable clustering, which can be compromised in dynamic environments with shifting client populations or evolving data characteristics, where computational complexity requires multi-objective optimization \cite{APAT2025103163, 10430628}.

\subsection{Personalized Federated Learning}

Personalized FL (PFL) addresses the challenge of local data heterogeneity by allowing each client to fine-tune a shared global model or maintain personalized components \cite{9743558,Zhang_Hua_Wang_Song_Xue_Ma_Guan_2023}. This improves local performance by adapting to individual client distributions. However, it differs from standard FL by focusing on client-specific optimization, which resolves local heterogeneity but does not offer a unified global solution.

\subsection{Optimization and Aggregation Enhancements}

This line of research aims to improve training efficiency and convergence under heterogeneous conditions by refining global optimization and aggregation strategies. FedOpt and its variants (FedAdam, FedYogi) \cite{reddi2021adaptivefederatedoptimization} apply adaptive learning rate techniques based on aggregated gradients, resulting in more stable updates across diverse clients. Additionally, hyperparameter optimization methods such as FedAdap \cite{10272438,10654271} fine-tune training parameters to improve performance in non-IID scenarios. While these techniques enhance optimization dynamics and support convergence, they do not fully resolve deeper issues related to data distribution skew, system variability, or long-term generalization in complex federated settings.

In summary, while existing federated learning methods have advanced the handling of non-IID data, they often struggle to effectively balance the influence of data quantity and distribution heterogeneity during aggregation. Many approaches address individual aspects like system heterogeneity, feature shifts, or client drift, yet a unified, principled solution that dynamically accounts for both dataset size and label distribution similarity remains lacking. To address this, we propose FedTVD, a novel aggregation method that leverages robust metrics to fairly and efficiently weight client contributions, leading to more stable, equitable, and generalized global models.

\begin{figure*}[!ht]
  \centering
   \includegraphics[width=0.99\textwidth]{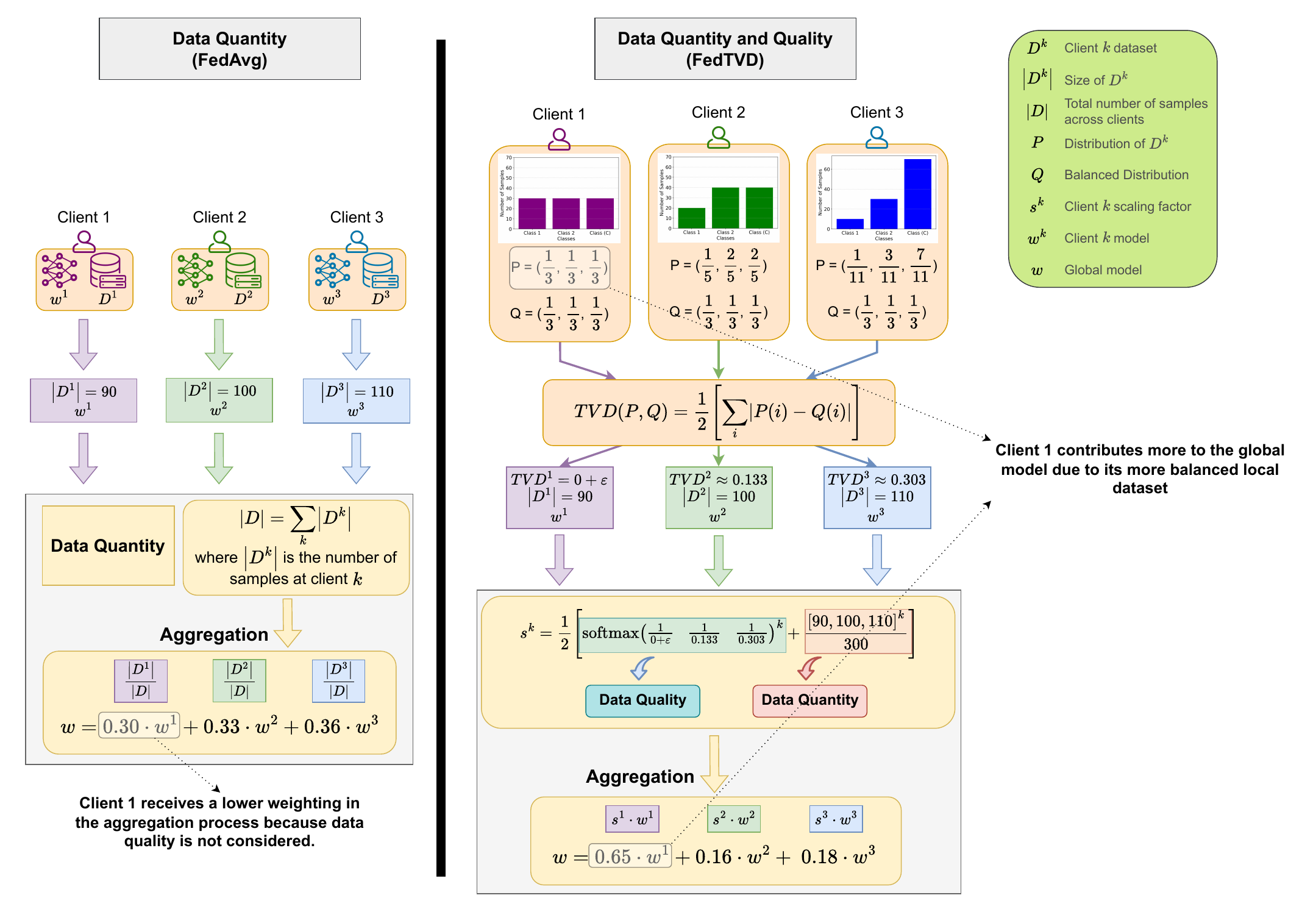}
    \caption{Comparison of FedAvg and FedTVD aggregation strategies. FedAvg relies only on data quantity (number of samples), while FedTVD incorporates both data quantity and quality, ensuring more balanced aggregation and mitigating model drift in non-IID settings.}
    \label{fedtvd_draw}
\end{figure*}

\section{Methodology}\label{sec3}

This section introduces the FedTVD algorithm and explains how it addresses the challenges of model aggregation in non-IID FL environments. We first discuss how the imbalance between data quantity and data quality in client updates can negatively affect model convergence and performance. We then present the FedTVD strategy, which corrects for this imbalance by jointly considering both factors during aggregation. Finally, we provide a detailed breakdown of the algorithm, including the client-server workflow and how TVD is incorporated into the aggregation process.

\subsection{Overview of FedTVD}

FedTVD is an aggregation strategy that considers both the size of each client's dataset (data quantity) and how well that dataset represents the global data distribution (data quality). This dual perspective enables more balanced and effective model aggregation unlike FedAvg, which relies solely on data quantity.

Data quantity refers to the total number of local samples available to each client. In conventional FL frameworks such as FedAvg, client aggregation weights are determined solely based on dataset size, under the assumption that larger datasets produce more reliable updates. However, in non-IID scenarios, larger datasets may be significantly biased toward specific classes and thus may not contribute proportionally to the generalization of the global model.

In contrast, data quality measures how closely a client's local data distribution aligns with the global model distribution, which is typically assumed to be IID and class-balanced.
For example, in a classification problem with three classes, the ideal global distribution would allocate one-third of the data to each class. In practical FL settings, clients typically exhibit varying degrees of deviation from this ideal distribution due to user-specific or environmental factors.
To quantify this deviation, we employ the TVD, a statistical metric that measures the disparity between a client’s local label distribution and the global balanced distribution. A TVD of zero indicates a perfectly balanced client dataset, while higher TVD values signal more severe distributional skewness.

Fig. \ref{fedtvd_draw} illustrates the difference between FedAvg and FedTVD across three clients. Each client holds samples from three classes. Client 1 has a balanced dataset with (30, 30, 30) samples per class, yielding a TVD of zero. Client 2 displays moderate imbalance with (20, 40, 40), and Client 3 has a notably skewed dataset of (10, 30, 70). In FedAvg, Client 3 receives the highest aggregation weight, driven solely by its larger dataset size. Conversely, FedTVD mitigates this issue by adjusting weights based on both data quantity and TVD. As a result, Client 1 receives a higher aggregation weight under FedTVD, reflecting its higher data quality, despite having fewer samples.
By dynamically balancing data quantity and quality, FedTVD promotes equitable and robust model aggregation, steering updates away from clients with skewed datasets and toward those with representative distributions. This fosters improved convergence and generalization in heterogeneous FL environments.

At the system level, FedTVD introduces minor yet impactful modifications to the standard FL workflow. Clients locally compute their label distributions and associated TVD values relative to the global distribution. These metrics, along with model updates and dataset sizes, are communicated to the server. The server then performs a refined aggregation by jointly incorporating data quantity and TVD, replacing the simplistic quantity-only weighting used in FedAvg.
The subsequent subsection provides a detailed description of the FedTVD workflow, outlining its client-server interactions and aggregation mechanism.

\subsection{Detailed FedTVD Algorithmic Steps}

The FedTVD algorithm follows the typical client-server paradigm in FL. In each round \( t \), the server distributes the current global model \( w_t \) to participating clients. Clients then update these parameters using their local datasets and send back the updated parameters, along with additional statistics. Below we provide a step-by-step outline of our approach, culminating in Algorithm \ref{fedTVD}. 

\subsubsection{Client-Side Operations}
\textbf{(1) Local Training:} Each client \( k \) begins with the global parameters \( w \) and proceeds to train on its private dataset \( D^k \) using an appropriate optimizer, such as stochastic gradient descent (SGD). Multiple local epochs can be executed if network constraints allow.

\textbf{(2) Label Distribution Computation:} Following training, the client computes its label distribution \( P \) by counting how many examples fall into each class. Formally, if the total number of classes is \( C \), the label distribution \( P \) is given by:
\begin{equation}
P = \left( \frac{|D^k_1|}{|D^k|}, \frac{|D^k_2|}{|D^k|}, \ldots, \frac{|D^k_C|}{|D^k|} \right)
\label{eq:label_dist}
\end{equation}
where \( |D^k_i| \) denotes the number of local samples of class \( i \), and \( |D^k| = \sum_{i=1}^{C} |D^k_i| \) is the total number of local samples for client \( k \).

To evaluate how this local distribution deviates from a balanced target, the client constructs a reference distribution \( Q \), in which all classes are assumed to be equally likely:
\begin{equation}
Q = \left( \frac{1}{C}, \frac{1}{C}, \ldots, \frac{1}{C} \right)
\label{eq:balanced_dist}
\end{equation}
This balanced distribution serves as a baseline for assessing the level of skew in the client’s local data.

\textbf{(3) TVD Calculation:} The client calculates the TVD to measure the deviation between its local data distribution \( P \) and the balanced reference distribution \( Q \). The TVD is defined as:
\begin{equation}
\text{TVD} = \frac{1}{2} \sum_{i=1}^{C} \left| P_i - Q_i \right|
\label{eq:tvd}
\end{equation}
TVD quantifies the total difference between the two distributions by summing the absolute differences of their corresponding probabilities. A higher TVD indicates a larger imbalance in the local data distribution, which is useful for identifying clients with more skewed data. This helps the server adjust the model aggregation process, giving less weight to clients with higher TVD values to ensure a more balanced global model.

\textbf{(4) Return of Results:} Finally, the client \( k \) sends its updated model parameters \( w^k \), the computed \( \text{TVD}^k \), and its local sample size \( |D^k| \) back to the server.

\subsubsection{Server-Side Operations}

\textbf{(1) Model Broadcast:} The server holds a global model parameter vector \( w \), which is initialized at the start. At each round \( t \), the server broadcasts \( w_t \) to all participating clients.

\textbf{(2) Reception of Client Updates:} Once clients finish their local training, the server collects the updated model parameters \( w^k \), \( \text{TVD}^k \), and local sample sizes \( |D^k| \) from each client \( k \) in the current round.

\textbf{(3) Softmax-Based TVD Weighting:} The server applies the softmax function to the set of TVD values received from all clients to transform them into a normalized weighting distribution:
\begin{equation}
a^k = \frac{ e^{-\text{TVD}^k} }{\sum_{j=1}^{K} e^{-\text{TVD}^j}}
\label{eq:softmax_tvd}
\end{equation}
This transformation adjusts each client's weight based on the degree of balance in their local label distribution. Clients with more balanced distributions (i.e., lower TVD values) are assigned higher weights, thereby contributing more significantly to the global model update. In contrast, clients with skewed distributions (higher TVD) are down-weighted. The softmax function also ensures a smooth and differentiable weighting scheme, preventing any single client from dominating the aggregation process.

\begin{algorithm}[!ht]
\caption{Federated Learning with TVD (FedTVD)}
\label{fedTVD}
\SetAlgoLined
\setlength{\abovedisplayskip}{0.1pt}
\setlength{\belowdisplayskip}{1.5pt}
\newcommand{\tightcomment}[1]{\tcp{#1}\vspace{-1.1em}}

\SetKwFunction{FMain}{CLIENT\_SIDE\_OPERATIONS}
\SetKwProg{Fn}{Function}{:}{}
\Fn{\FMain{$w$}}{
    Train local model $w$ on local data $D^k$ \\
    \tightcomment{Calculate local label distribution}
    \[
    P = \left( \frac{|D^k_1|}{|D^k|}, \frac{|D^k_2|}{|D^k|}, \ldots, \frac{|D^k_C|}{|D^k|} \right)
    \tag*{Eq. (\ref{eq:label_dist})}
    \] \\
    \tightcomment{Initialize global balanced distribution}
    \[
    Q = \left( \frac{1}{C}, \frac{1}{C}, \ldots, \frac{1}{C} \right)
    \tag*{Eq. (\ref{eq:balanced_dist})}
    \] \\
    \tightcomment{Calculate TVD}
    \[
    \text{TVD} = \frac{1}{2} \sum_{i=1}^{C} \left| P_i - Q_i \right|
    \tag*{Eq. (\ref{eq:tvd})}
    \] \\
    \Return $w$, $\text{TVD}$, and $|D^k|$
}
\textbf{End Function}

\SetKwFunction{FMain}{SERVER\_SIDE\_OPERATIONS}
\SetKwProg{Fn}{Function}{:}{}
\Fn{\FMain{}}{
    Initialize global model parameters $w$ \\
    \For{each round $t = 1, 2, \ldots, T$}{
        Distribute global model $w_t$ to all participating clients \\
        \For{each client $k = 1, 2, \ldots, K$}{
            $w^k$, $\text{TVD}^k$, and $|D^k| \gets \textit{CLIENT\_SIDE\_OPERATIONS}(w_t)$ \\
        }
        \tightcomment{Calculate softmax of TVD values}
        \[
        a^k = \frac{ e^{-\text{TVD}^k} }{\sum_{j=1}^{K} e^{-\text{TVD}^j}}
        \tag*{Eq. (\ref{eq:softmax_tvd})}
        \] \\
        \tightcomment{Calculate total number of samples across clients}
        \[
        |D| = \sum_{k=1}^{K} |D^k|
        \tag*{Eq. (\ref{eq:total_samples})}
        \] \\
        \tightcomment{Compute scaling factor}
        \[
        s^k = \lambda \cdot a^k + (1 - \lambda) \cdot \frac{|D^k|}{|D|}
        \tag*{Eq. (\ref{eq:scaling_factors})}
        \] \\
        \tightcomment{Aggregate using}
        \[
        w_{t+1} = \sum_{k=1}^{K} s^k \, w^k
        \tag*{Eq. (\ref{eq:weighted_avg})}
        \] \\
    }
}
\textbf{End Function}

\end{algorithm}

\textbf{(4) Composite Scaling Factors:} 
Following the softmax-based weighting described above, the server integrates each client’s weight \( a^k \), which reflects the balance of its local label distribution, with its relative data volume. First, it computes the total number of samples across all clients:
\begin{equation}
|D| = \sum_{k=1}^{K} |D^k|
\label{eq:total_samples}
\end{equation}
This allows the computation of the proportion of data held by each client \( k \), expressed as \( \frac{|D^k|}{|D|} \), which captures the contribution of the client in terms of dataset size.

The final scaling factor \( s^k \) for client \( k \) is then computed by combining the softmax-derived weight \( a^k \), which reflects the balance of its local label distribution, with its relative data volume. This combination is controlled via a tunable parameter \( \lambda \in [0, 1] \) as follows:
\begin{equation}
s^k = \lambda \cdot a^k + (1 - \lambda) \cdot \frac{|D^k|}{|D|}
\label{eq:scaling_factors}
\end{equation}
Here, \( \lambda \) enables flexible control over the influence of data quality versus quantity in the aggregation process. A higher \( \lambda \) emphasizes data quality (i.e., label balance), while a lower \( \lambda \) emphasizes dataset size. Notably, setting \( \lambda = 1 \) results in weighting purely based on data quality, and setting \( \lambda = 0 \) reduces the algorithm to FedAvg, where only dataset size is considered.

This composite scaling factor therefore accounts for both the quality of the client’s data (via \( a^k \), which penalizes distributional skew) and its quantity (via \( \frac{|D^k|}{|D|} \)). By jointly considering these two dimensions, the algorithm ensures that clients with both large and well-balanced datasets exert greater influence on the global model, while clients with limited or highly skewed data are appropriately down-weighted. This strategy enables a principled trade-off between fairness and statistical representativeness in federated model aggregation.

\textbf{(5) Model Aggregation:} Finally, the server aggregates the local models into the new global model \( w_{t+1} \) using:
\begin{equation}
w_{t+1} = \sum_{k=1}^{K} s^k \, w^k
\label{eq:weighted_avg}
\end{equation}
This updated model \( w_{t+1} \) is then broadcast to the clients in the next round.

\subsection{Complexity Analysis}
To evaluate the scalability of FedTVD, we analyze its computational complexity and compare it to that of FedAvg. We consider the dominant operations per communication round on both client and server sides.

\textbf{Client-Side Complexity:} The primary computational cost on the client side is local training. Assuming each client performs \( E \) epochs of SGD over their local dataset \( D^k \), the training complexity is:
\[
\mathcal{O}(E \cdot |D^k| \cdot |w|)
\]
where \( |w| \) is the number of model parameters. In addition to training, FedTVD computes the local label distribution and TVD. Label counting has linear complexity in the dataset size and class count, \( \mathcal{O}(|D^k| + C) \), while TVD computation is \( \mathcal{O}(C) \). Given that \( C \) is typically much smaller than \( |D^k| \), the total client-side complexity remains dominated by the training phase.


\textbf{Server-Side Complexity:} At the server, FedAvg aggregates \( K \) client models per round, costing \( \mathcal{O}(K \cdot |w|) \). FedTVD introduces lightweight additional operations:
\begin{itemize}
    \item Softmax over \( K \) TVD values: \( \mathcal{O}(K) \)
    \item Total sample count: \( \mathcal{O}(K) \)
    \item Composite scaling factor computation: \( \mathcal{O}(K) \)
    \item Weighted model aggregation: \( \mathcal{O}(K \cdot |w|) \)
\end{itemize}

Although FedTVD adds four extra steps at the server, all but one are lightweight \( \mathcal{O}(K) \) scalar operations. Only the final step—weighted model aggregation—shares the same dominant complexity \( \mathcal{O}(K \cdot |w|) \) as FedAvg. Thus, FedTVD preserves the same overall order of server-side complexity as FedAvg, with only minor constant-factor overhead.

\subsection{Convergence Analysis}

We analyze the convergence behavior of FedTVD under standard assumptions in federated optimization. Our approach builds upon the convergence theory of FedAvg, which we briefly recall here to provide a foundation for comparison.

In FedAvg, the global objective function is defined as:
\[
F(w) = \sum_{i=1}^N p_i F_i(w), \quad \text{where} \quad p_i = \frac{|D_i|}{\sum_{j=1}^N |D_j|},
\]
and each local objective function \( F_i(w) \) is assumed to be \( L \)-smooth and \( \mu \)-strongly convex. Under these assumptions, FedAvg achieves the following convergence bound after \( T \) communication rounds:
\[
\mathbb{E}[F(w^T) - F(w^*)] \leq O\left(\frac{1}{T}\right) + O(\delta_{\text{FedAvg}}^2),
\]
where all gradients are evaluated at the current global model \( w^t \), and the client drift term is defined as:
\[
\delta_{\text{FedAvg}}^2 = \frac{1}{K} \sum_{k=1}^K \left\| \nabla F_k(w^t) - \nabla F(w^t) \right\|^2.
\]

FedTVD modifies only the aggregation step of FedAvg. Instead of relying solely on the sample proportion \( |D_i| \), it incorporates both data quantity and data quality. In particular, each client is assigned a weight \( s^k \) that depends on its dataset size and the TVD of its label distribution.

The updated global model is then computed using a normalized weighted average:
\[
w^{t+1} = \sum_{k=1}^K s^k w_k^t, \quad \text{with} \quad \sum_{k=1}^K s^k = 1, \quad s^k \geq 0.
\]
This structure remains consistent with the class of algorithms analyzed in FedAvg, and thus FedTVD inherits the same theoretical convergence guarantees under convexity and smoothness.

Formally, the expected optimality gap for FedTVD satisfies:
\[
\mathbb{E}[F(w^T) - F(w^*)] \leq O\left(\frac{1}{T}\right) + O(\delta_{\text{FedTVD}}^2),
\]
where the modified client drift term is given by:
\[
\delta_{\text{FedTVD}}^2 = \sum_{k=1}^K s^k \left\| \nabla F_k(w^t) - \nabla F(w^t) \right\|^2.
\]

Since TVD captures the degree of label imbalance, which often leads to biased local gradients, FedTVD naturally reduces drift by assigning lower weights to clients with highly skewed distributions. This weighting effect suppresses the contribution of highly biased updates, which implies that \( \delta_{\text{FedTVD}}^2 \leq \delta_{\text{FedAvg}}^2 \) in typical non-IID scenarios.

In conclusion, FedTVD maintains the same convergence rate as FedAvg under standard convexity and smoothness assumptions. By incorporating distributional information into the aggregation process, it reduces the impact of client drift, which is often the dominant source of error in non-IID FL. As a result, FedTVD is theoretically as sound as FedAvg while offering improved robustness in practical scenarios.

\subsection{Advantages and Novelty}
\noindent \textbf{(A) Mitigating non-IID Effects:} 
FedTVD explicitly moderates clients exhibiting large distributional discrepancies from a balanced reference distribution. By computing the TVD of each client’s local label distribution with respect to an ideal (uniform) distribution, our method identifies those with skewed or biased data. This mechanism prevents the global model from becoming dominated by a few skewed distributions, ensuring a more equitable update that remains resilient in heterogeneous settings.

\vspace{1mm}
\noindent \textbf{(B) Robustness to Label Skew Outliers:}
Clients with highly imbalanced label distributions can act as statistical outliers and introduce bias during model aggregation. FedTVD addresses this issue by assigning lower aggregation weights to such clients using an exponential penalty based on their TVD. This reduces their influence on the global model and helps improve robustness against distributional skew. We note that our notion of robustness is limited to mitigating the effect of label imbalance, and does not address other types of outliers such as clients with corrupted data labels or adversarial clients \cite{10274102}.

\vspace{1mm}
\noindent \textbf{(C) Balanced Data-Driven Weighting:}
While addressing skewness is vital, large datasets may hold significant information. FedTVD reconciles these needs by incorporating both the TVD-based weight (\( a^k \)) and the sample-size fraction (\( \frac{|D^k|}{|D|} \)). This hybrid scheme strikes a balance between distributional alignment and data volume, preventing large but skewed datasets from overwhelming the global model while still leveraging their information-rich contribution.

\vspace{1mm}
\noindent \textbf{(D) Light Overhead:}
Compared to traditional FL, the main overhead of FedTVD is the computation and transmission of TVD for each client. However, this overhead remains minimal compared to raw data exchange or large-scale gradient sharing. The primary computational overhead in FedTVD arises from computing the TVD metric, which involves summing over class distributions, This complexity is negligible compared to the overall cost of FL, making FedTVD scalable to large deployments.

\vspace{1mm}
\noindent \textbf{(E) Extensibility:}
Though introduced in the context of a standard FL approach, the TVD modification is designed as a lightweight, drop-in mechanism that can be integrated into diverse federated algorithms. Because it involves minimal additional computation and does not require substantial changes to communication or optimization routines, it can be readily adopted in any FL framework to handle non-IID data.

\vspace{1mm}
\noindent \textbf{(F) Real-World Use Cases:}
FedTVD is particularly well-suited to FL environments where client data is skewed due to user-specific, demographic, or operational factors. In healthcare applications, for instance, different hospitals or medical centers may collect patient data that naturally varies in disease prevalence or age distribution, which leads to highly imbalanced local datasets. FedTVD can mitigate the influence of such skewed clients during model aggregation while preserving meaningful contributions. Similarly, in large-scale sensor networks or environmental monitoring, data collected from different regions or sensor types may reflect localized patterns that deviate from the overall distribution. While FedTVD computes TVD using a uniform reference distribution to avoid domination by any particular class, this reference can be adapted in practice to reflect domain-specific priors if a more accurate global distribution is known. This flexibility ensures that FedTVD remains applicable across both standard and application-aware FL scenarios, enhancing fairness and robustness in aggregation.

\vspace{2mm}
\noindent \textbf{Summary:} 
In conclusion, FedTVD offers a balanced and adaptive aggregation strategy by coupling data quantity with a principled TVD-based penalty. This approach effectively addresses heterogeneous data distributions and outlier risks while imposing minimal overhead. The key operations and line-by-line equations are outlined in Algorithm \ref{fedTVD}, providing a clear blueprint for practical deployment in a wide range of FL scenarios.

\section{Experimental Setup}\label{sec4}

\begin{table*}[!b]
\centering
\caption{Mean test accuracy for the last 10 rounds across different datasets (FMNIST, CIFAR-10, and CIFAR-100) under varying Dirichlet non-IID distributions (\( \alpha = 0.1, 0.5, 1.0 \)) and IID settings.}
\label{exp1}
\begin{tblr}{
  cells = {c},
  cell{2}{1} = {r=4}{},
  cell{2}{2} = {r=4}{},
  cell{6}{1} = {r=4}{},
  cell{6}{2} = {r=4}{},
  cell{10}{1} = {r=4}{},
  cell{10}{2} = {r=4}{},
  hline{1,14} = {-}{0.08em},
  hline{2,6,10} = {-}{0.05em},
}
\textbf{Dataset} & \textbf{Model} & \textbf{Dir ($ \alpha $)} & \textbf{FedAvg} & \textbf{FedProx} & \textbf{FedNova} & \textbf{FedDkw} & \textbf{FedTVD (ours)}\\
FMNIST & CNN & 0.1 & 82.64 ± 0.71 & 80.69 ± 1.68 & 83.63 ± 0.22 & 83.99 ± 0.61 & \textbf{85.26 ± 0.65}\\
 &  & 0.5 & 86.11 ± 0.56 & 86.08 ± 0.39 & 86.65 ± 0.14 & 87.20 ± 0.19 & \textbf{87.21 ± 0.21}\\
 &  & 1.0 & 87.69 ± 0.36 & 87.62 ± 0.41 & 88.01 ± 0.10 & 88.03 ± 0.21 & \textbf{88.39 ± 0.14}\\
 &  & IID & \textbf{88.95 ± 0.16} & 88.83 ± 0.25 & 88.89 ± 0.15 & 88.89 ± 0.25 & 88.74 ± 0.23\\
CIFAR10 & Resnet-18 & 0.1 & 48.24 ± 3.35 & 49.44 ± 2.69 & 53.52 ± 2.20 & 56.30 ± 1.61 & \textbf{58.82 ± 1.75}\\
 &  & 0.5 & 73.33 ± 0.42 & 72.61 ± 1.00 & 74.26 ± 0.19 & 74.51 ± 0.37 & \textbf{74.69 ± 0.29}\\
 &  & 1.0 & 76.17 ± 0.32 & 75.22 ± 0.62 & 75.37 ± 0.43 & 75.13 ± 0.41 & \textbf{76.22 ± 0.45}\\
 &  & IID & 77.56 ± 0.37 & 77.34 ± 0.45 & 77.23 ± 0.32 & 77.12 ± 0.40 & \textbf{78.04 ± 0.25}\\
CIFAR100 & Resnet-34 & 0.1 & 35.31 ± 1.12 & 36.12 ± 0.44 & 36.26 ± 0.57 & 36.16 ± 0.87 & \textbf{37.68 ± 0.38}\\
 &  & 0.5 & 38.91 ± 0.36 & 38.36 ± 0.75 & 38.89 ± 0.44 & 38.21 ± 1.08 & \textbf{39.58 ± 0.71}\\
 &  & 1.0 & 39.30 ± 0.55 & 39.30 ± 0.92 & 39.09 ± 0.27 & 39.06 ± 0.69 & \textbf{39.79 ± 0.51}\\
 &  & IID & 39.24 ± 0.68 & 39.33 ± 0.29 & 39.76 ± 0.59 & 39.95 ± 0.20 & \textbf{40.51 ± 0.45}
\end{tblr}
\end{table*}

This section outlines the experimental setup, which includes the datasets, data partitioning techniques, computational environment, and configurations used to evaluate the proposed approach. In order to evaluate the model's effectiveness under various client participation rates and data heterogeneity levels, we provide a comprehensive understanding of its performance in diverse scenarios.

\subsection{Implementation Details and System Configuration}  
All experiments were conducted using Python 3.11, with PyTorch 2.2.2 serving as the core deep learning framework. FL simulations were implemented using the FedEasy framework \cite{fedeasy}, enabling efficient orchestration of client-server interactions. To ensure consistency across experimental runs, all computations were performed on a high-performance computing server equipped with an Intel Core i9-10900X CPU running at 3.70 GHz, 192 GB of RAM, and four NVIDIA GeForce RTX 4090 GPUs. This computational setup facilitated large-scale federated training while maintaining reproducibility and efficiency.

\subsection{Datasets and Model Architectures}
The evaluation was conducted on three standard benchmark datasets: FashionMNIST (FMNIST) \cite{xiao2017fashion}, CIFAR-10 \cite{krizhevsky2009learning}, and CIFAR-100, each presenting a different level of complexity. FMNIST, a grayscale image classification dataset, was used with a standard Convolutional Neural Network (CNN) architecture \cite{lecun1998gradient}. CIFAR-10 was evaluated using a ResNet-18 model \cite{he2015deepresiduallearningimage}, while CIFAR-100 was trained on a deeper ResNet-34 architecture. These datasets provided a diverse range of classification challenges, allowing for an in-depth analysis of the proposed method’s robustness across both simple and complex tasks.

\subsection{Data Partitioning and Experimental Variants}
To simulate realistic non-IID conditions, data partitioning was performed using a Dirichlet allocation strategy \cite{hsu2019measuringeffectsnonidenticaldata}, controlling the extent of statistical heterogeneity across clients. The concentration parameter \(\alpha\) was varied to reflect different levels of non-IID data distributions, with lower values \(\alpha = 0.1\) indicating highly skewed distributions and higher values \(\alpha = 1\) representing more balanced partitions. For reference, a fully IID scenario was also included in the analysis by setting \(\alpha\) to a sufficiently large value.

Two major experimental settings were designed to analyze the impact of data heterogeneity and CPR. The first set of experiments investigated the effect of varying non-IID levels on model convergence and performance across all three datasets. In this setting, different values of \(\alpha = 0.1, 0.5, and 1\) were considered, allowing an in-depth evaluation of the impact of statistical heterogeneity. Each experiment involved 100 clients, with 10\% of them randomly selected in each round to participate in training.
The second set of experiments specifically focused on assessing how the number of clients and CPR influence learning dynamics. These trials were conducted using the CIFAR-10 dataset under a highly non-IID setting \(\alpha = 0.1\). The number of clients was varied between 100 and 200, while the CPR was systematically adjusted to 10\%, 20\%, and 50\%, enabling an analysis of scalability and participation effects on model convergence.

\subsection{Hyperparameters and Training Process}
To ensure fair comparisons, all experiments followed a standardized training protocol. Model updates were optimized using SGD with a learning rate of 0.01 and a momentum coefficient of 0.9. Each client performed four local epochs per training round with a batch size of 32 before transmitting model updates to the central server. Training was conducted over 300 communication rounds to thoroughly evaluate model convergence and stability across different settings. For FedTVD, the hyperparameter \( \lambda \) was fixed at 0.50, reflecting an equal balance between client label distribution quality and dataset size during aggregation. To ensure robustness and reproducibility, each experiment was repeated five times using different random seeds, with data resampling performed independently in each experiment for training only. The testing dataset was fixed on the server and remained unchanged throughout all runs. The reported results represent the mean \(\pm\) standard deviation over these runs.

\begin{figure*}[!ht]
    \centering
    \begin{subfigure}{0.32\textwidth}
        \includegraphics[width=\textwidth]{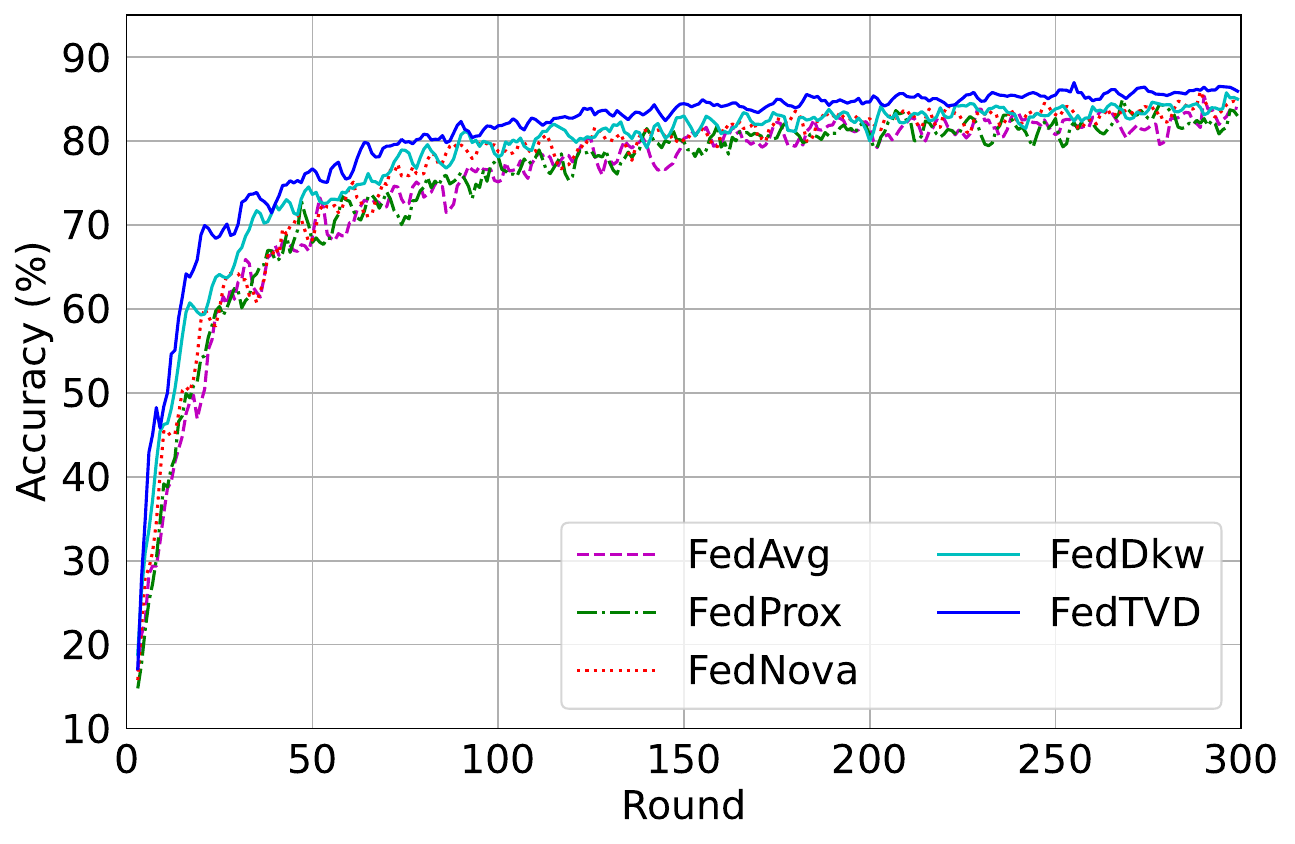}
        \caption{\( \alpha = 0.1 \)}
        \label{fig:fmnist_subfig1}
    \end{subfigure}
    \hfill
    \begin{subfigure}{0.32\textwidth}
        \includegraphics[width=\textwidth]{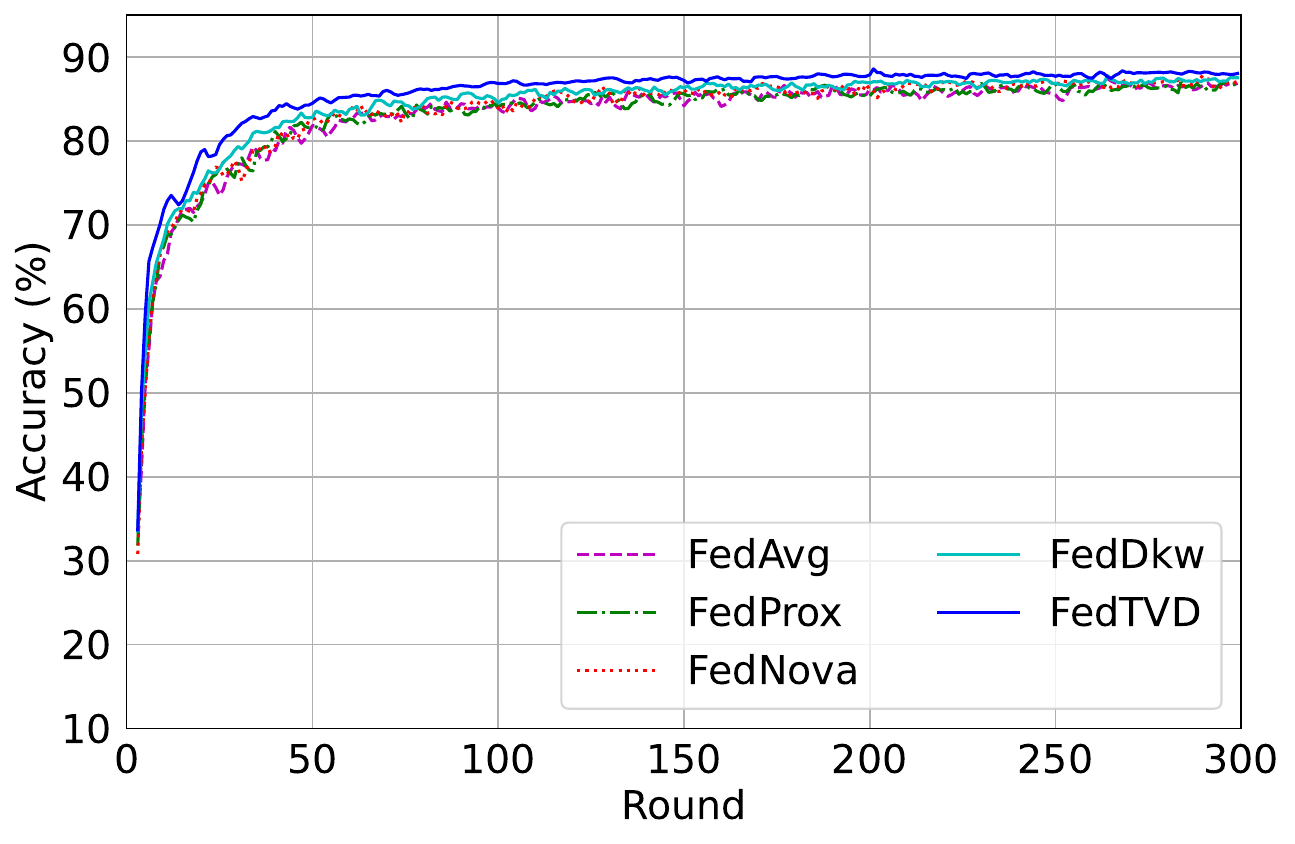}
        \caption{\( \alpha = 0.5 \)}
        \label{fig:fmnist_subfig2}
    \end{subfigure}
    \hfill
    \begin{subfigure}{0.32\textwidth}
        \includegraphics[width=\textwidth]{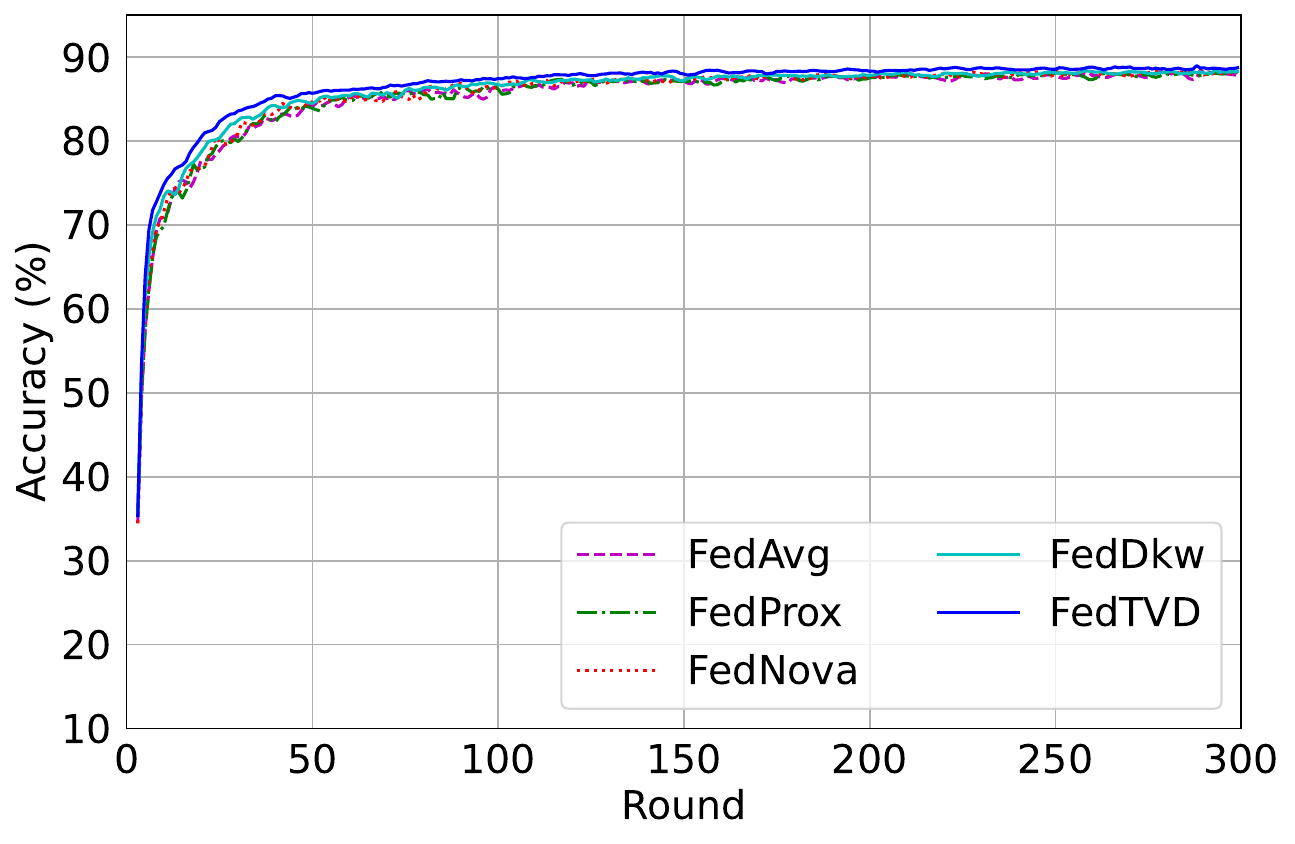}
        \caption{\( \alpha = 1.0 \)}
        \label{fig:fmnist_subfig3}
    \end{subfigure}
    \caption{Accuracy trends over rounds for the FMNIST dataset using a CNN model architecture under different levels of non-IID data distribution controlled by the Dirichlet parameter \( \alpha \).}
    \label{fig:fmnist}
\end{figure*}
\begin{figure*}[!ht]
    \centering
    \begin{subfigure}{0.32\textwidth}
        \includegraphics[width=\textwidth]{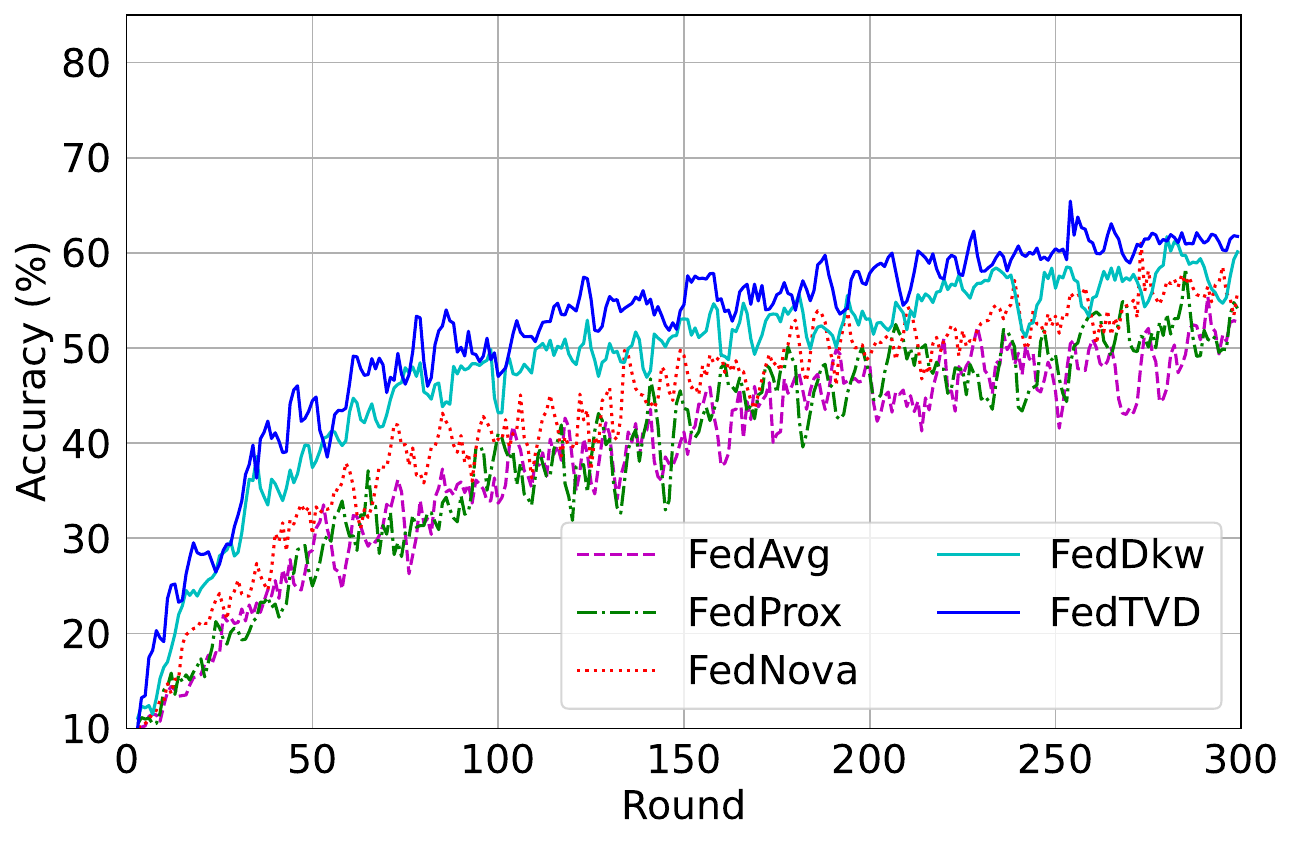}
        \caption{\( \alpha = 0.1 \)}
        \label{fig:cifar10_subfig1}
    \end{subfigure}
    \hfill
    \begin{subfigure}{0.32\textwidth}
        \includegraphics[width=\textwidth]{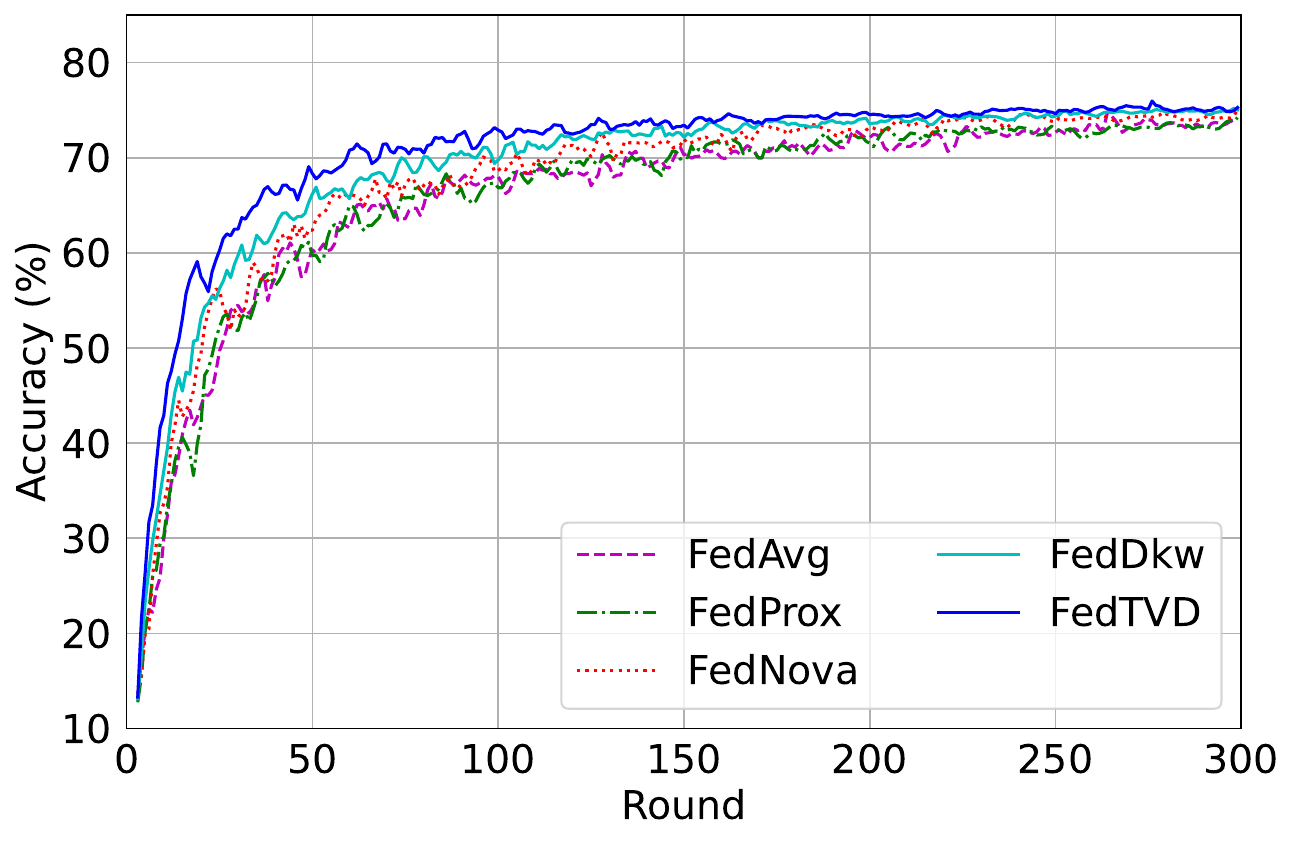}
        \caption{\( \alpha = 0.5 \)}
        \label{fig:cifar10_subfig2}
    \end{subfigure}
    \hfill
    \begin{subfigure}{0.32\textwidth}
        \includegraphics[width=\textwidth]{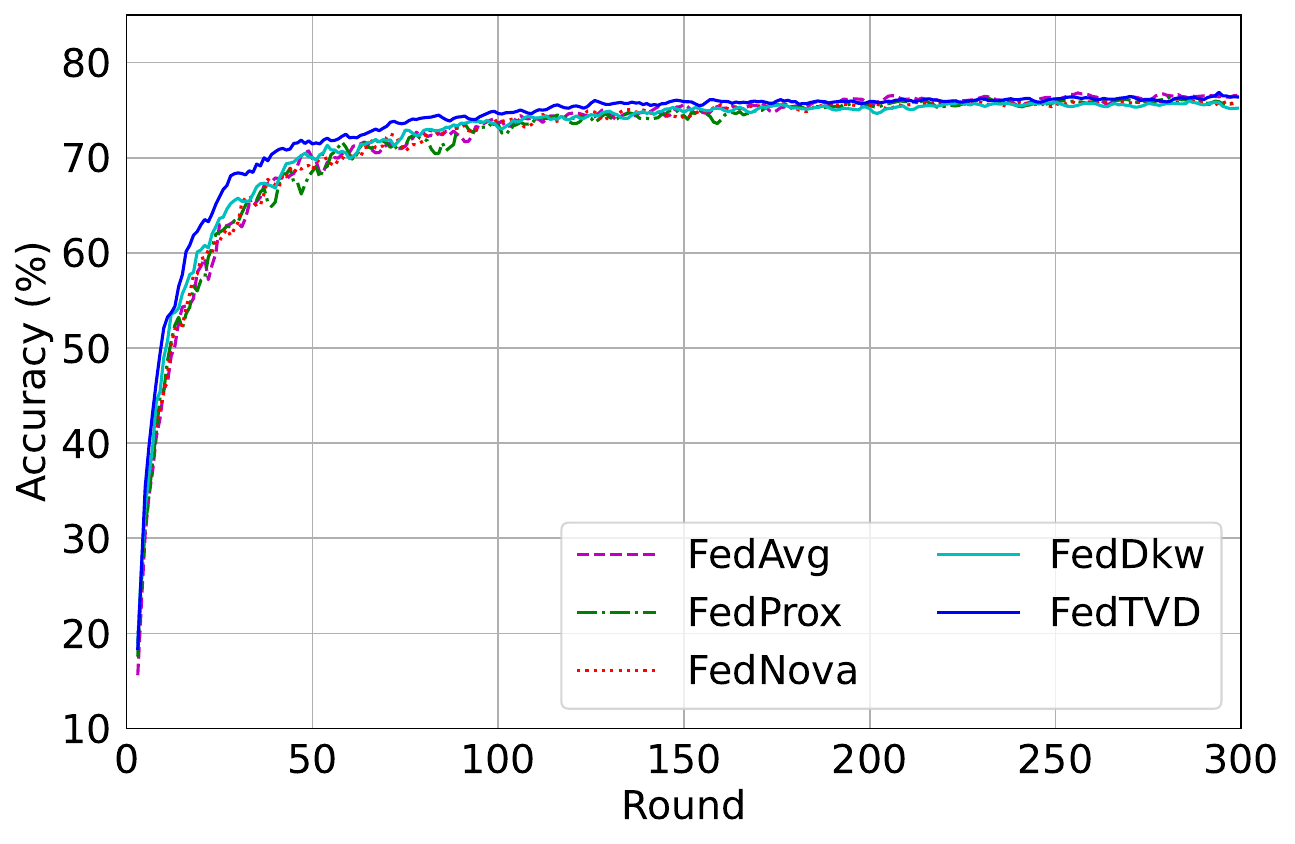}
        \caption{\( \alpha = 1.0 \)}
        \label{fig:cifar10_subfig3}
    \end{subfigure}
    \caption{Accuracy trends over rounds for the CIFAR-10 dataset using a ResNet-18 model architecture under different levels of non-IID data distribution controlled by the Dirichlet parameter \( \alpha \).}
    \label{fig:cifar10}
\end{figure*}
\begin{figure*}[!ht]
    \centering
    \begin{subfigure}{0.32\textwidth}
        \includegraphics[width=\textwidth]{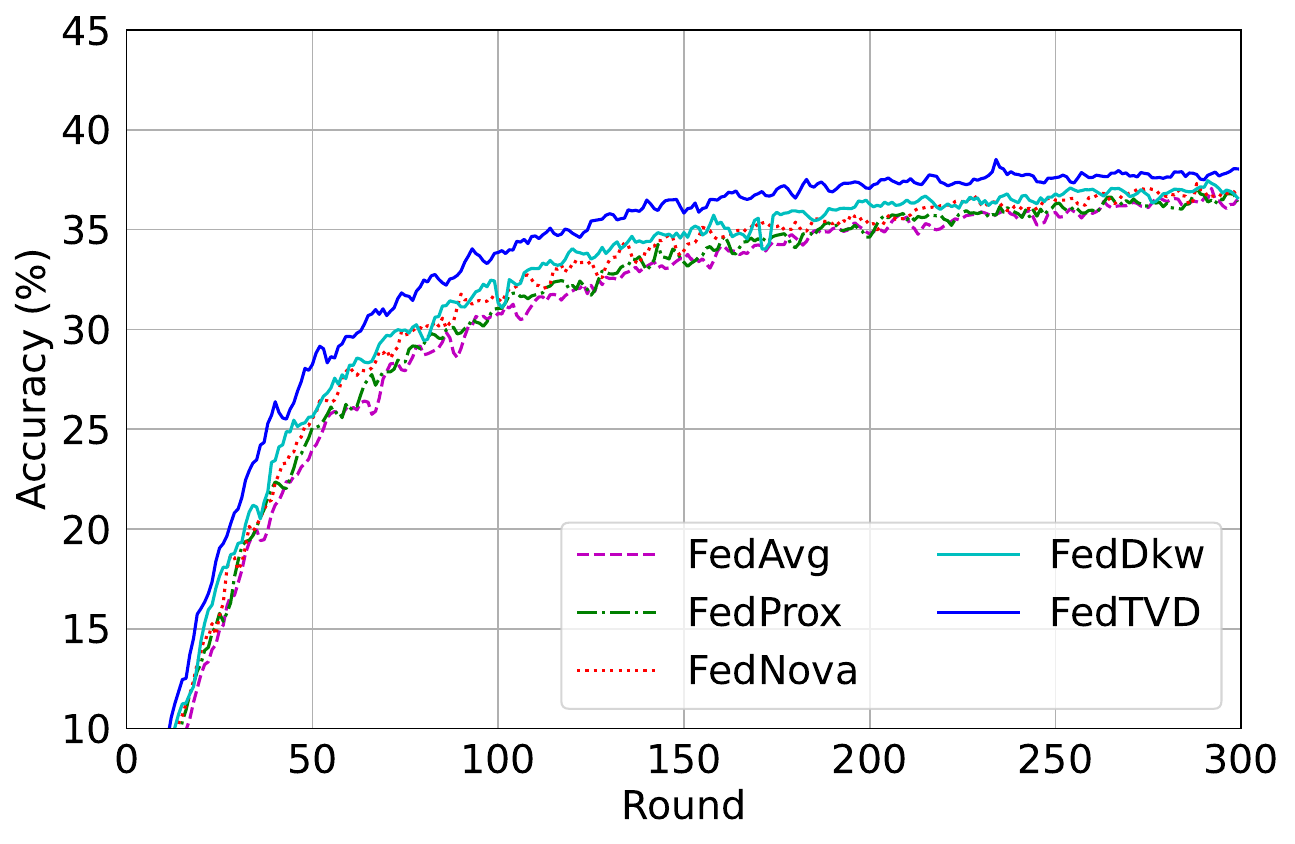}
        \caption{\( \alpha = 0.1 \)}
        \label{fig:cifar100_subfig1}
    \end{subfigure}
    \hfill
    \begin{subfigure}{0.32\textwidth}
        \includegraphics[width=\textwidth]{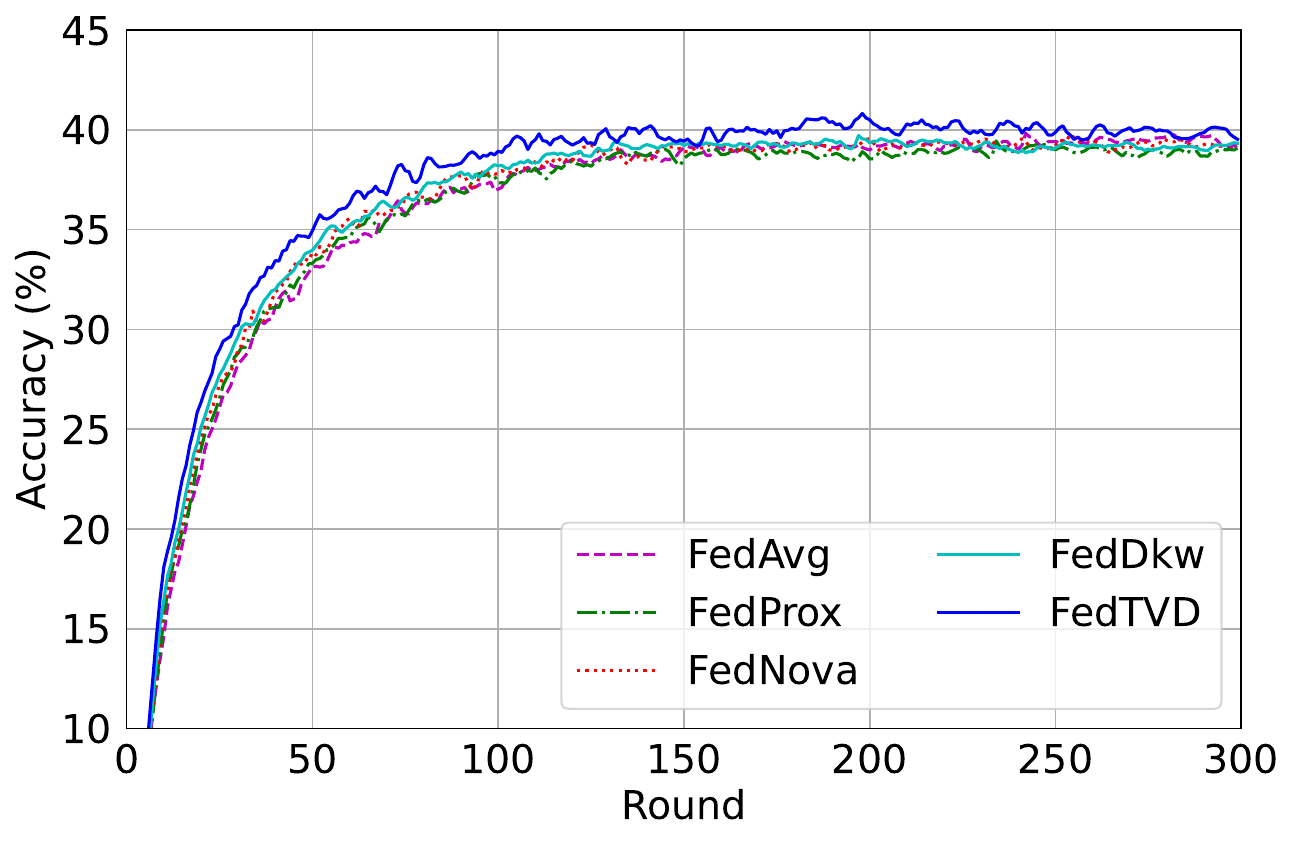}
        \caption{\( \alpha = 0.5 \)}
        \label{fig:cifar100_subfig2}
    \end{subfigure}
    \hfill
    \begin{subfigure}{0.32\textwidth}
        \includegraphics[width=\textwidth]{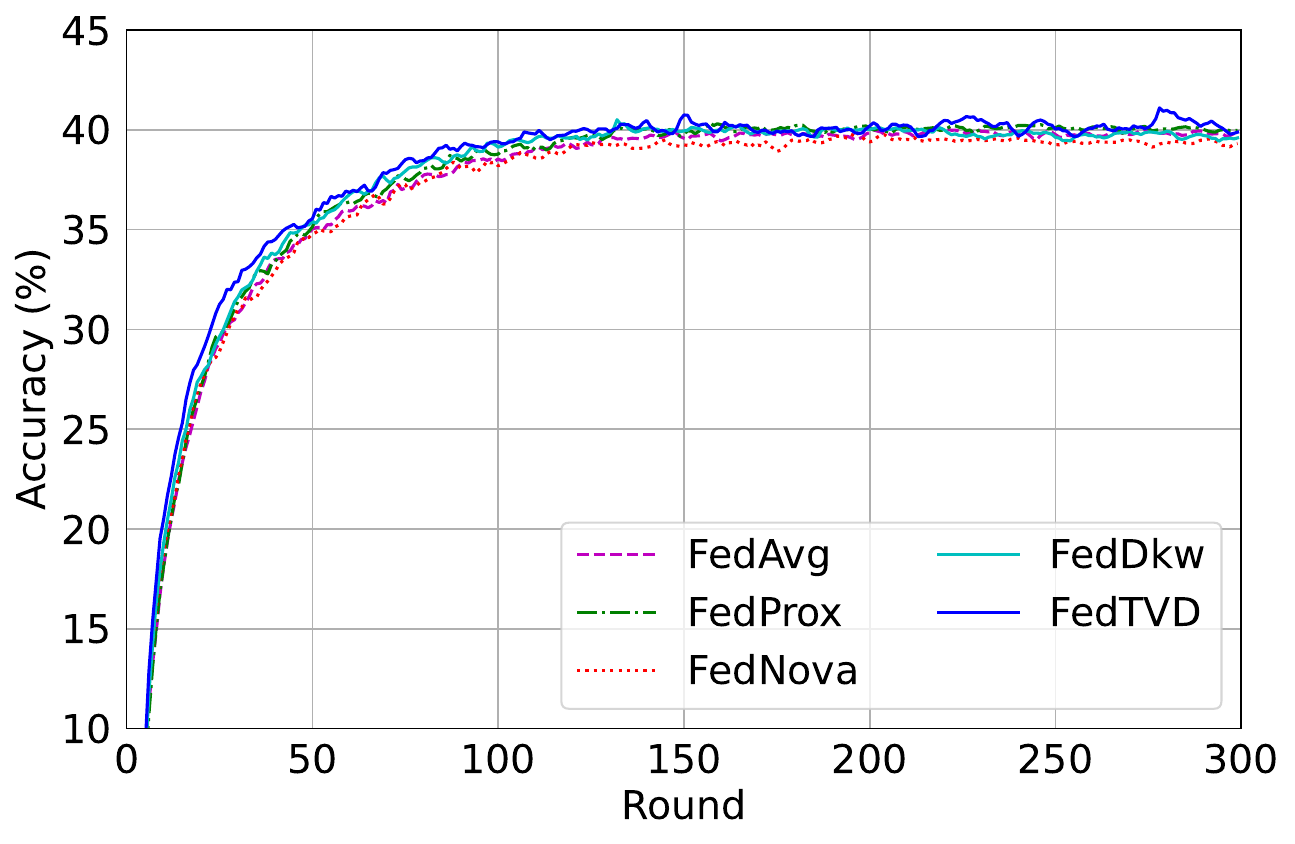}
        \caption{\( \alpha = 1.0 \)}
        \label{fig:cifar100_subfig3}
    \end{subfigure}
    \caption{Accuracy trends over rounds for the CIFAR-100 dataset using a ResNet-34 model architecture under different levels of non-IID data distribution controlled by the Dirichlet parameter \( \alpha \).}
    \label{fig:cifar100}
\end{figure*}

\section{Results \& Discussion}\label{sec5}

\begin{table*}[!ht]
\centering
\caption{Mean test accuracy for the last 10 rounds on the CIFAR-10 dataset under extreme non-IID settings (\( \alpha = 0.1 \)) with varying numbers of clients and CPR.}
\label{exp2}
\begin{tblr}{
  cells = {c},
  cell{2}{1} = {r=3}{},
  cell{5}{1} = {r=3}{},
  hline{1,8} = {-}{0.08em},
  hline{2,5} = {-}{0.05em},
}
\textbf{\#Clients} & \textbf{CPR} & \textbf{FedAvg} & \textbf{FedProx} & \textbf{FedNova} & \textbf{FedDkw} & \textbf{FedTVD (ours)}\\
100 & 0.1 & 48.24 ± 3.35 & 49.44 ± 2.69 & 53.52 ± 2.20 & 56.30 ± 1.61 & \textbf{58.82 ± 1.75}\\
 & 0.2 & 56.45 ± 0.82 & 56.55 ± 1.58 & 58.37 ± 0.77 & 59.98 ± 1.14 & \textbf{61.68 ± 1.57}\\
 & 0.5 & 60.60 ± 0.94 & 61.18 ± 0.82 & 62.00 ± 0.65 & 62.35 ± 0.89 & \textbf{62.93 ± 0.86}\\
200 & 0.1 & 46.71 ± 1.04 & 45.28 ± 2.42 & 52.91 ± 1.84 & 55.11 ± 1.83 & \textbf{57.39 ± 3.30}\\
 & 0.2 & 52.05 ± 2.49 & 50.99 ± 2.22 & 56.24 ± 1.48 & 57.57 ± 1.61 & \textbf{58.78 ± 1.88}\\
 & 0.5 & 55.18 ± 2.26 & 56.41 ± 2.17 & 58.62 ± 1.49 & 59.19 ± 1.18 & \textbf{61.12 ± 0.97}
\end{tblr}
\end{table*}

In this section, we present the results from comprehensive comparative experiments conducted to verify the effectiveness of the proposed FedTVD algorithm. These experiments benchmark FedTVD against several state-of-the-art FL algorithms, including FedAvg, FedProx, FedNova, and FedDkw. The results, presented in the following subsections, demonstrate the superior performance of FedTVD in terms of test accuracy and convergence speed, particularly in extreme non-IID data settings.

\subsection{Performance Analysis Across Varying Data Distributions and Datasets}

In scenarios with extreme data heterogeneity, where the Dirichlet parameter \( \alpha = 0.1 \), FedTVD demonstrates a clear advantage over other algorithms. As shown in Fig. \ref{fig:fmnist_subfig1}, \ref{fig:cifar10_subfig1} and Fig. \ref{fig:cifar100_subfig1}, FedTVD achieves faster convergence and higher accuracy compared to FedAvg, FedProx, and FedNova, particularly on the CIFAR-10 dataset. The results in Table \ref{exp1} also reinforce this finding, where FedTVD reaches 58.82\% test accuracy on CIFAR-10, surpassing FedAvg (48.24\%) and FedDkw (56.30\%). This performance is especially significant in environments where the clients' data are highly skewed, resulting in significant heterogeneity between client datasets. In such settings, FedTVD's adaptive aggregation mechanism prioritizes the updates that contribute most effectively to improving the global model, which helps mitigate the negative impact of extreme data imbalances. Specifically, FedTVD is able to focus on the more informative updates from clients with data distributions that are more representative of the global model, thus enabling faster and more stable convergence. The benefit of FedTVD becomes more pronounced as the data heterogeneity increases, demonstrating its ability to adapt to highly non-IID settings and outperform other methods that struggle to converge efficiently under such conditions. In conclusion, FedTVD excels in highly heterogeneous environments, where its adaptive aggregation scheme allows it to outperform other FL algorithms by better handling the skewed data distributions across clients.

As data heterogeneity becomes more moderate, represented by \( \alpha = 0.5 \) and \( \alpha = 1.0 \), FedTVD maintains its advantage in both accuracy and convergence speed. The trends in Fig. \ref{fig:cifar10_subfig2}, Fig. \ref{fig:cifar10_subfig3}, Fig. \ref{fig:cifar100_subfig2}, and Fig. \ref{fig:cifar100_subfig3}, along with Table \ref{exp1}, demonstrate that FedTVD consistently outperforms FedAvg, FedProx, and FedNova. On CIFAR-100, FedTVD achieves 39.58\% accuracy for \( \alpha = 0.5 \), slightly surpassing FedAvg (38.91\%) and FedDkw (38.21\%). For \( \alpha = 1.0 \), it reaches 39.79\%, outperforming FedAvg (39.30\%) and FedDkw (39.06\%). This suggests that even in less extreme non-IID scenarios, FedTVD effectively balances client contributions and mitigates data imbalance, leading to stable and efficient convergence. Unlike FedAvg and FedNova, which struggle with slower adaptation in moderately heterogeneous settings, FedTVD dynamically adjusts client weights to prioritize more informative updates. This adaptability results in consistent performance improvements across both CIFAR-10 and CIFAR-100.

Even in IID settings, where client data distributions are uniform, FedTVD maintains competitive accuracy and stable convergence, as shown in Table \ref{exp1} and Figs. \ref{fig:fmnist_subfig3} and \ref{fig:cifar10_subfig3}. While the performance gap with existing methods like FedAvg and FedDkw is marginal, FedTVD's adaptive aggregation remains effective, ensuring consistent updates across clients. Its ability to perform well across both IID and non-IID environments highlights its versatility in FL. This minor gain may arise because, despite IID sampling, the Dirichlet-based partitioning with finite \( \alpha \)  can still lead to slight class imbalances, allowing the TVD term to act as a weak regularizer.

\subsection{Influence of Total Number of Clients and CPR}

This section analyzes the performance according to CPR with 100 clients. For a low CPR of 0.1, where only 10\% of clients participate in each communication round, FedTVD achieves an accuracy of 58.82\%, outperforming FedDkw (56.30\%) and FedAvg (48.24\%), as shown in Table \ref{exp2}. As CPR increases, FedTVD's performance improves, reaching 62.93\% at a CPR of 0.5, maintaining its lead over FedDkw (62.35\%). This improvement can be attributed to the increased diversity and quantity of updates as more clients contribute to the aggregation process, allowing FedTVD to better capture the global data distribution and enhance test set accuracy. FedTVD benefits from broader data access, further boosting its performance.

As the client count increases to 200, the overall performance of all algorithms slightly decreases due to greater data distribution heterogeneity and reduced data volume per client. At a CPR of 0.1, FedTVD achieves 57.39\%, slightly outperforming FedDkw (55.11\%) and significantly surpassing FedAvg (46.71\%). At a higher CPR of 0.5, FedTVD achieves 61.12\%, maintaining its lead over FedDkw (59.19\%). The reduced accuracy compared to the 100-client scenario is expected, as the broader distribution of data increases variability in local updates, and the reduced data per client limits the representativeness of individual models. Despite these challenges, FedTVD’s robust adaptive mechanism minimizes the performance drop by aggregating the most relevant updates, ensuring that the global model quality remains high even with a larger number of clients.
\vfill

\subsection{Impact of Data Quality}

In FL, both data quantity and data quality influence model performance. This section explores the impact of quality-aware weighting in FL aggregation by comparing FedAvg (which prioritizes dataset size) and FedTVD (which dynamically adjusts weights based on both label distribution and dataset size) on the CIFAR-10 dataset under a Dirichlet distribution with $\alpha = 0.1$, using 100 clients (CPR of 0.1).

\begin{figure*}
    \centering
    \begin{subfigure}{0.48\textwidth}
        \includegraphics[width=\linewidth]{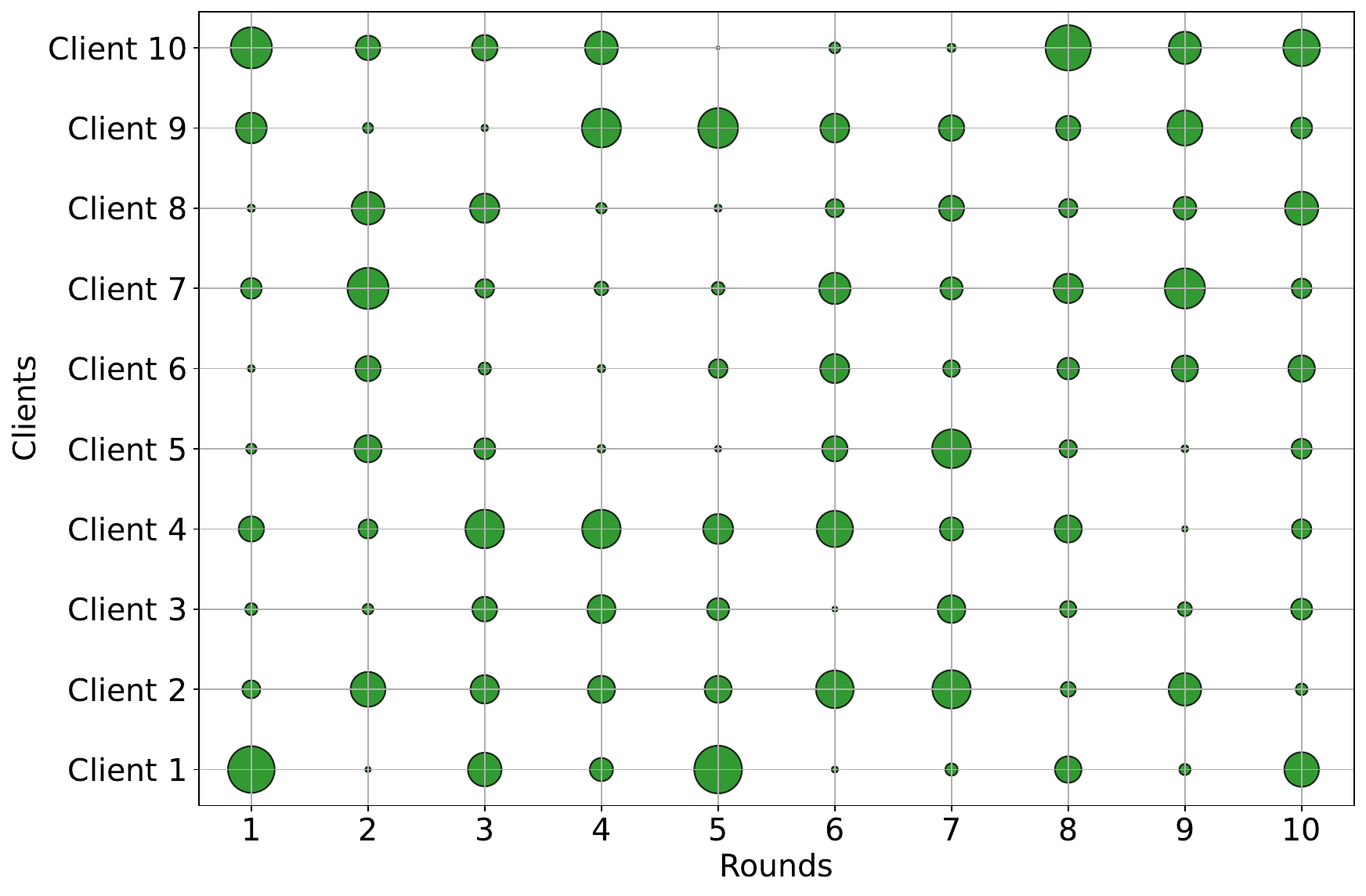}
        \caption{Data Quantity only}
        \label{fig:fedavg_only}
        \vspace{5pt}
    \end{subfigure}
    \begin{subfigure}{0.48\textwidth}
        \includegraphics[width=\linewidth]{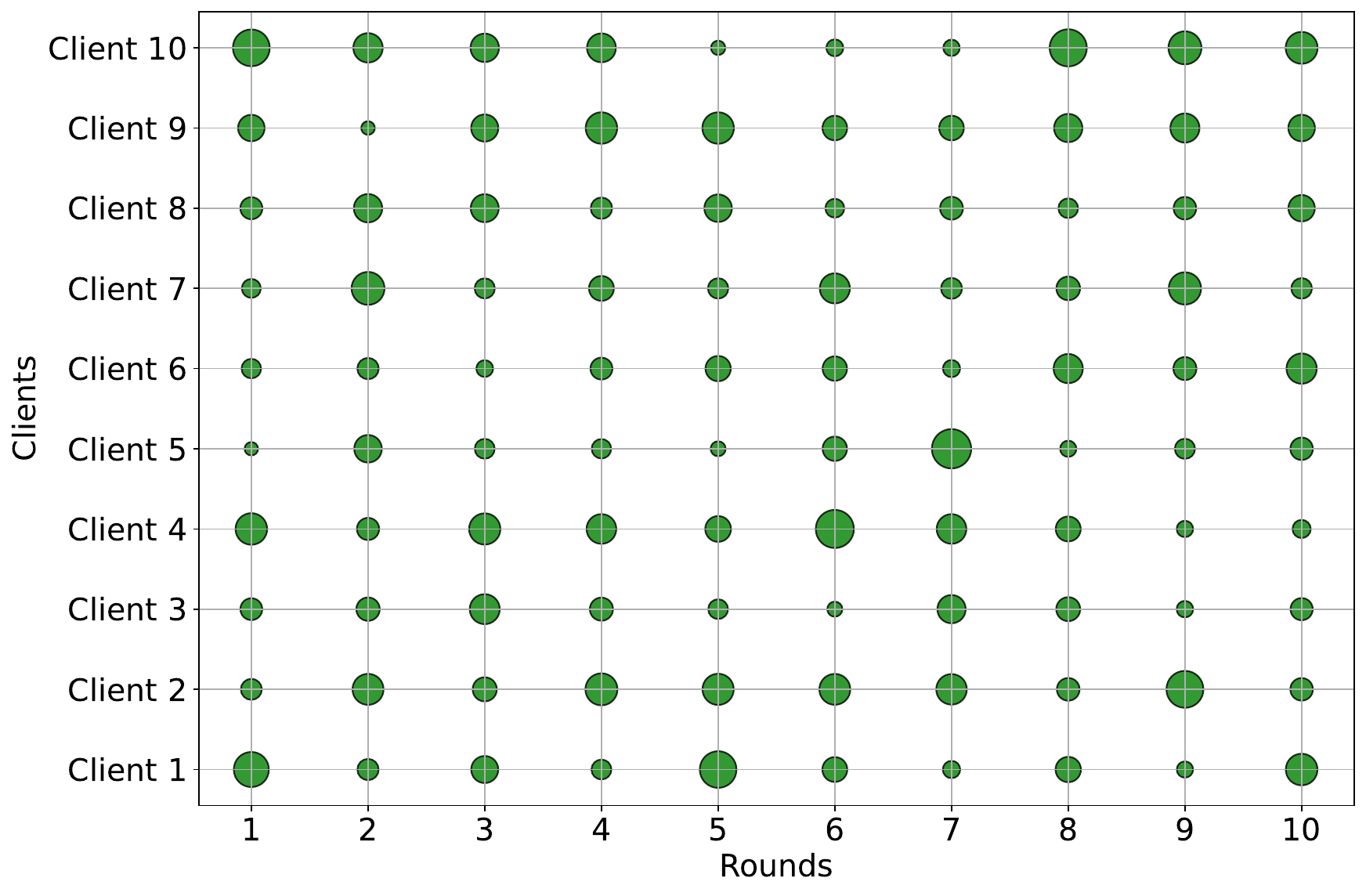}
        \caption{Data Quantity and Quality}
        \label{fig:fedavg_and_fedtvd}
    \end{subfigure}
    \caption{Conceptual illustration of different weighting strategies in FL on the CIFAR-10 dataset under a Dirichlet distribution with $\alpha = 0.1$, using 100 clients (CPR of 0.1). The bubble size represents the relative contribution of each client during aggregation. (a) When only data quantity is considered, clients with larger datasets dominate the updates. (b) A combined approach balances both factors, ensuring that high-quality contributions are prioritized alongside larger datasets.}
    \label{fig:data_quantity_quality}
\end{figure*}

Fig. \ref{fig:data_quantity_quality} illustrates the difference between traditional quantity-based aggregation (Fig. \ref{fig:fedavg_only}) and FedTVD’s quality-aware approach (Fig. \ref{fig:fedavg_and_fedtvd}). In Fig. \ref{fig:fedavg_only}, larger datasets dominate aggregation, leading to an imbalance where a few clients contribute excessively while others have minimal influence. In contrast, Fig. \ref{fig:fedavg_and_fedtvd} shows a more even distribution of contributions, as FedTVD dynamically adjusts weights based on both dataset size and label distribution. This ensures that clients with highly skewed data do not disproportionately impact the global model while still leveraging large, high-quality datasets. As a result, FedTVD achieves a 10.58\% accuracy improvement over FedAvg in extreme non-IID settings ($\alpha = 0.1$, Table \ref{exp1}), demonstrating the importance of incorporating data quality into FL aggregation.

\captionsetup{labelfont=bf,labelsep=newline, hypcap=false}
\vspace{10pt}
\begin{minipage}{0.9\columnwidth}
\centering
\small
\captionof{table}{Mean test accuracy over the last 10 rounds on CIFAR-10 for different \( \lambda \) values under two non-IID settings: extreme (\( \alpha = 0.1 \)) and moderate (\( \alpha = 1.0 \)). Each setting uses 100                                                                                               clients (CPR of 0.1).}
\label{lambda_ablation}
\begin{tabular}{c|c|c}
\toprule
\multirow{2}{*}{\textbf{\( \lambda \)}} & \multicolumn{2}{c}{\textbf{Accuracy (mean ± std)}} \\
\cmidrule{2-3}
& \( \alpha = 0.1 \) & \( \alpha = 1.0 \) \\
\midrule
0.00 & 48.24 ± 3.35 & 76.17 ± 0.32 \\
0.25 & 55.88 ± 0.71 & \textbf{76.25 ± 0.22} \\
0.50 & \underline{58.82 ± 1.75} & \underline{76.22 ± 0.45} \\
0.75 & 58.16 ± 0.79 & 76.05 ± 0.55 \\
1.00 & \textbf{60.18 ± 0.96} & 76.03 ± 0.52 \\
\bottomrule
\end{tabular}
\end{minipage}
\vspace{10pt}

To further examine the impact of balancing data quantity and quality, we provide an ablation study on the weighting parameter $\lambda$, which controls the trade-off between the two components during aggregation. A value of $\lambda = 0.00$ corresponds to using data quantity only (i.e., FedAvg), while $\lambda = 1.00$ corresponds to FedTVD's full reliance on data quality. Intermediate values interpolate between these extremes.

As shown in Table \ref{lambda_ablation}, under extreme non-IID conditions (\( \alpha = 0.1 \)), even small increases in \( \lambda \) lead to notable accuracy gains—for example, \( \lambda = 0.25 \) improves performance by over 7\% compared to \( \lambda = 0.00 \). Although \( \lambda = 1.00 \) achieves the highest accuracy in this extreme setting, it performs slightly worse in the moderate skewed setting (\( \alpha = 1.0 \)). Relying solely on data quality, as with \( \lambda = 1.00 \), may not generalize well across all FL scenarios since, in less skewed environments, data quantity becomes more important. In contrast, \( \lambda = 0.50 \) delivers consistently strong results across both scenarios—ranking second-best for \( \alpha = 0.1 \) and matching the top accuracy for \( \alpha = 1.0 \). This balanced approach, which combines contributions from both data quality and quantity, makes \( \lambda = 0.50 \) a robust default choice for diverse FL environments.

These results highlight the critical role of balancing data quality and quantity in FL aggregation. Unlike traditional methods that overweight large but potentially biased datasets, FedTVD with \( \lambda = 0.50 \) ensures a fairer contribution from all clients, leading to better generalization across diverse non-IID scenarios and more stable convergence.
\vfill

\section{Conclusion}\label{sec6}

FedTVD advances the state of the art in FL by incorporating a lightweight yet effective metric, the TVD, into the aggregation procedure. By combining TVD-derived weights with sample-based proportions, it captures both distributional alignment and data volume, helping to mitigate the common challenges of non-IID data. This leads to more robust convergence, particularly in heterogeneous settings. FedTVD's ability to handle outliers and imbalanced data makes it well-suited for real-world applications where client data is naturally diverse.

While our approach specifically targets label distribution skew, it does not directly address feature distribution skew, concept drift, or fairness across subgroups defined by non-label features. Tackling these more complex forms of heterogeneity would require significant methodological extensions and richer datasets. We view this as an important future direction, with potential to further improve fairness, robustness, and adaptability in large-scale federated systems.




\section*{Funding}
This research was supported by Basic Science Research Program through the National Research Foundation of Korea (NRF) funded by the Ministry of Education (No. NRF-2022R1I1A3072355).

\section*{Data Availability}
The data used in this study are publicly available.

\section*{Declaration of Competing Interests}
The authors declare that they have no known competing financial interests or personal relationships that could have appeared to
influence the work reported in this paper.

\printcredits

\bibliographystyle{unsrtnat}

\bibliography{cas-refs}



\end{document}